\documentclass[lettersize,journal]{IEEEtran}
\usepackage{amsmath,amsfonts}
\usepackage{algorithmic}
\usepackage{tabularx}
\usepackage{algorithm}
\usepackage{array}
\usepackage[caption=false,font=normalsize,labelfont=sf,textfont=sf]{subfig}
\usepackage{textcomp}
\usepackage[dvipsnames]{xcolor}
\usepackage{stfloats}
\usepackage{url}
\usepackage{verbatim}
\usepackage{graphicx}
\usepackage{cite}
\usepackage{tikz}

\usepackage{fancyhdr}

\fancypagestyle{titlepage}{
  \fancyhf{}
  \fancyhead[L]{\scriptsize This article has been accepted for publication in IEEE Robotics \& Automation Magazine. This is the author's version which has not been fully edited and content may change prior to final publication. Citation information: DOI 10.1109/MRA.2024.3415108}
  \fancyfoot[C]{\scriptsize \textcopyright \the\year{} IEEE. Personal use is permitted, but republication/redistribution requires IEEE permission. See https://www.ieee.org/publications/rights/index.html for more information.}
}

\begin{document}

\title{An Interactive Augmented Reality Interface\\ for Personalized Proxemics Modeling}

\author{Massimiliano Nigro, Amy O'Connell, Thomas Groechel, Anna-Maria Velentza and Maja Matari\'c,~\IEEEmembership{Fellow,~IEEE}

\thanks{Massimiliano Nigro, Amy O'Connell, Thomas Groechel, Anna-Maria Velentza and Maja Matari\'c are all
with the Interaction Lab, Department of Computer Science,
University of Southern California, Los Angeles, CA 90089, USA}
}


\maketitle
\thispagestyle{titlepage}

\begin{abstract}

Understanding and respecting personal space preferences is essential for socially assistive robots designed for older adult users. This work introduces and evaluates a novel personalized context-aware method for modeling users' proxemics preferences during human-robot interactions. Using an interactive augmented reality interface, we collected a set of user-preferred distances from the robot and employed an active transfer learning approach to fine-tune a specialized deep learning model. We evaluated this approach through two user studies: 1) a convenience population study (N = 24) to validate the efficacy of the active transfer learning approach; and 2) a user study involving older adults (N = 15) to assess the system's usability. We compared the data collected with the augmented reality interface and with the physical robot to examine the relationship between proxemics preferences for a virtual robot versus a physically embodied robot. We found that fine-tuning significantly improved model performance: on average, the error in testing decreased by 26.97\% after fine-tuning. The system was well-received by older adult participants, who provided valuable feedback and suggestions for future work.
\end{abstract}


\section{Introduction}

\IEEEPARstart{O}{lder} adults value their independence but also often experience loneliness and social isolation \cite{national2020social}. Over the past two decades, researchers have explored using in-home socially assistive robots (SARs) to provide companionship to older adults. Several aspects of robot behavior have emerged as critical factors affecting the adoption of these systems.

 Lack of trust in the robot and perceived technology complexity are two such factors that negatively impact SAR acceptance \cite{pino2015we}. How a robot shares space with and maintains an appropriate distance from the user, referred to as \textit{proxemic behavior}, influences its perceived sociability and trustworthiness. Perceived enjoyment, sociability, and adaptivity positively influence SAR acceptance among older adults \cite{pino2015we}. Taking these factors into account, approaches to forming proxemics models for SARs should be easy and enjoyable for the user, and should adapt to their unique needs and preferences through continued interactions.

Respecting personal space is a highly complex social skill, influenced by a multitude of factors such as user age \cite{okita2012captain} and cultural context \cite{eresha2013investigating}. The robot's physical appearance \cite{bhagya2019exploratory} and position in the environment \cite{yasumoto2011personal,walters2007robot} also affect the range of appropriate robot distances during a human-robot interaction (HRI).

\begin{figure}[!hptb]
    \centering   \includegraphics[width=0.45\textwidth]{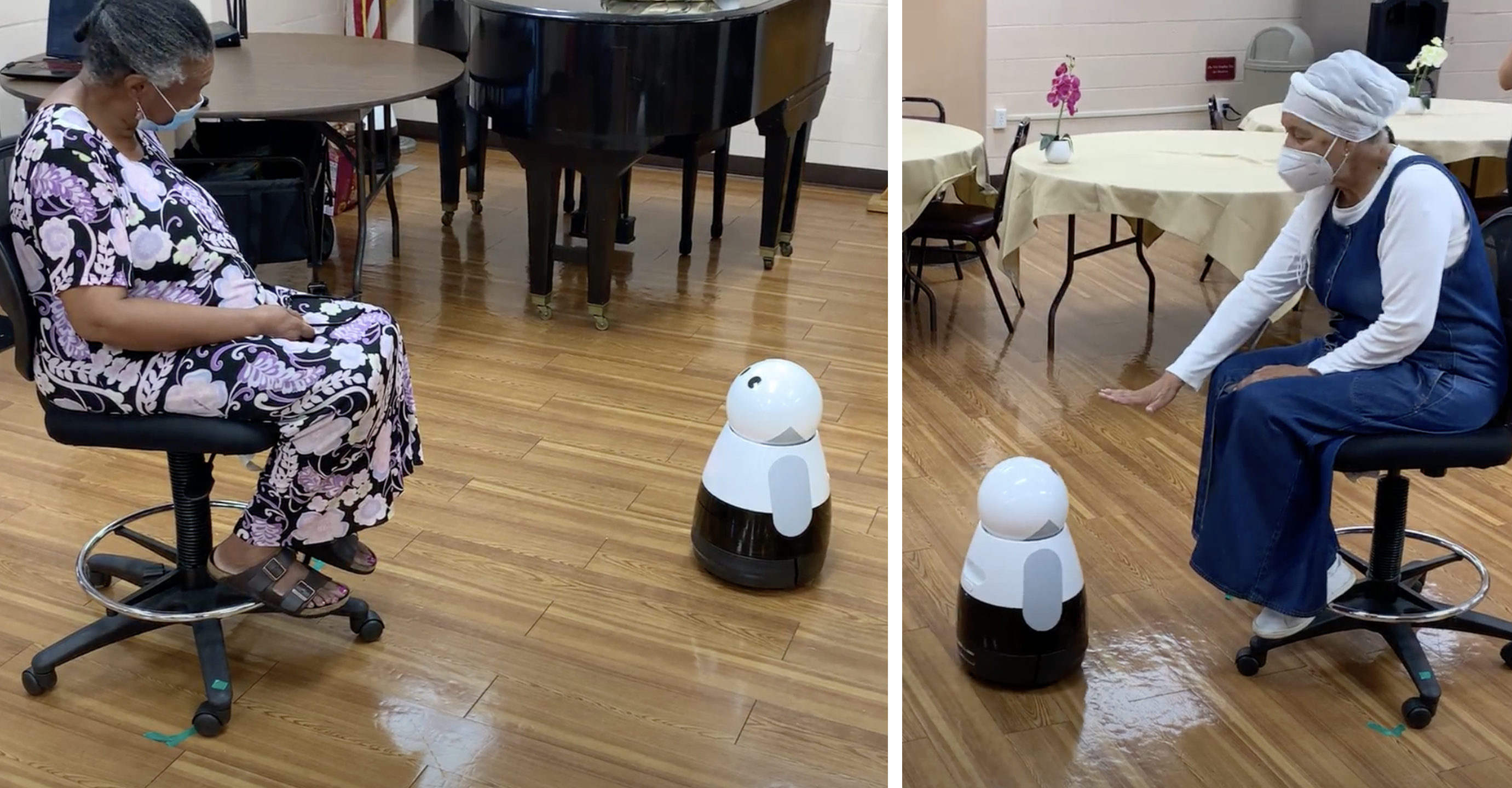}
    \caption{Older adults have a variety of proxemics preferences}
    \label{fig:physical_kuri}
\end{figure}

\begin{figure}[!hptb]
    \centering   \includegraphics[width=0.45\textwidth]{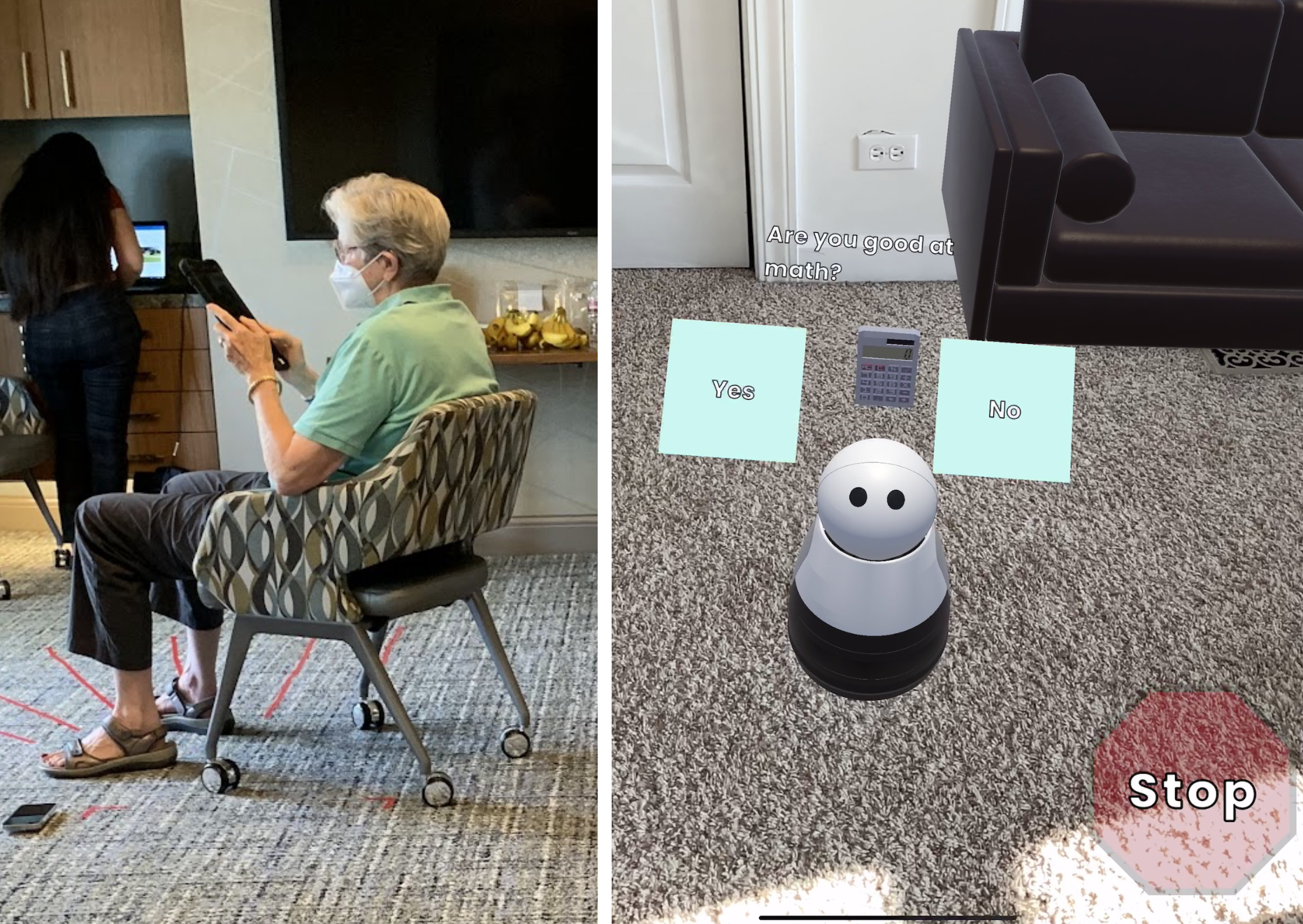}
    \caption{Study participants used an AR interface to indicate when the approaching virtual robot was at a comfortable distance}
    \label{fig:virtual_kuri}
\end{figure}

Most current approaches to collecting user proxemics preferences are either unable to produce personalized proxemics models of individual users or too inflexible or time-consuming to be repeated in different environments.

Gao et al.  \cite{gao_investigating_2018} explored deep learning approaches to predict a user's discomfort based on the robot's relative position and the user's demographic data. The training data were collected in a motion capture studio using reflective markers to track the robot and user, an approach that would be difficult to replicate in other settings. 

Patompak et al. \cite{patompak_learning_2020} proposed an approach in which the parameters of a fuzzy inference system were optimized through reinforcement learning to form a personalized social interaction model. In their study, a robot interacted with the participants from different positions in a room, forming personalized maps of user discomfort based on the participants' responses. This approach enabled personalized proxemics modeling through real-world interactions between the robot and user. However, interactions that involve a physical robot traveling to different positions in the environment are limited by the robot's physical constraints. For example, the user may spend several seconds between interactions waiting for the robot to navigate to the next sampled location.

Our work circumvents this issue by utilizing a virtual robot visualized in the user's physical environment through an augmented reality (AR) interface. We leverage the capabilities of AR to design a question-answering conversation activity that is engaging for the user, and we compare the set of proxemics preferences collected in this activity to preferences gathered under a second experimental condition with a physical robot to understand the relationship between user proxemics preferences with a virtual robot versus a physical robot. We use the collected data and an active transfer learning (ATL) \cite{monarch2021human} approach to form personalized proxemics models for each user.

In summary, this work contributes the following:
\begin{itemize}
    \item A method for collecting user proxemics preference data via an engaging interaction with a virtual agent through an AR interface;
    \item An approach to personalizing deep learning proxemics models for individual users through ATL.
\end{itemize}

In the sections that follow, we introduce our technical approach to proxemics modeling and the system we have designed for this purpose. We then describe two user studies we conducted to 1) evaluate our ATL approach to generating personalized proxemics models; and 2) validate our AR interface and activity for usability with a sample of older adult users. 

\section{Technical Approach}

In this section, we introduce the problem formulation, the models used to predict the user’s proxemics preferences, and the design of the system and AR application.

\subsection{Problem Formulation}

\begin{figure}[!hptb]
    \centering
    \includegraphics[width=0.3\textwidth]{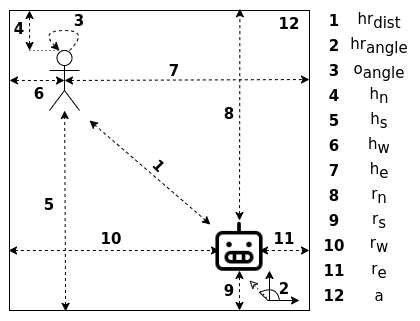}
    \caption{Example of features considered in this work}
    \label{fig:features}
\end{figure}

\label{problem_formulation}

Following Gao et al. and Patompak et al. \cite{gao_investigating_2018, patompak_learning_2020}, we sought to predict user discomfort in an HRI scenario defined by the following features:

\begin{itemize}
    \item $hr_{dist}$, the distance between the user and the robot;
    \item $hr_{sin}$,$hr_{cos}$, the angle between the user and the robot;
    \item $o_{sin}, o_{cos}$, the orientation of the user relative to the robot;
    \item $h_n,h_s,h_w,h_e$, the distances between the user and the closest environment boundaries (e.g., walls) in the cardinal directions;
    \item $r_n,r_s,r_w,r_e$, the distances between the robot and the closest environment boundaries in the cardinal directions;
    \item $a$, the area of the environment.
\end{itemize}
These features are depicted in Figure \ref{fig:features}. We represent user discomfort as an ordinal score between 0 and 100. We only consider scenarios involving one stationary user and one approaching mobile robot in the environment.

Because our objective is to form personalized models that learn the proxemics preferences of individual users, we approach the problem as a {\it domain adaptation problem} \cite{zhuang2020comprehensive}, where our objective is to adapt the predictions of a model trained on a general pool of data, a \textit{training} domain, to an \textit{application} domain, a pool of data relevant to a specific user. 

\subsection{Proxemics Model}\label{generalmodel}
We trained our proxemics model on the public dataset SocNav1 \cite{manso2020socnav1} that contains a collection of 8168 synthetic HRI scenarios in which the appropriateness of the robot's position is rated on a scale from 0 to 100. Each HRI scenario in the dataset is characterized by: a scenario identifier, a robot identifier, a list of points defining the walls, a list of humans, a list of objects, a list of interaction tuples (either human-human or human-object pairs), and the robot position rating described above.
We filtered the data to only include scenarios with a single human who isn't interacting with any objects is in the room.
After filtering, 1300 training instances remained in the training set. Given the limited training instances for a regression problem, we chose to frame the problem as an ordinal regression problem and to use a feedforward neural network architecture. We also applied a smoothing function to the predictions during the testing phase. We chose to use the Savitzky-Golay smoothing filter \cite{press1990savitzky} and experimentally determined that models predicting soft ordinal labels \cite{diaz2019soft}, rather than ordinal classes, performed better. Following D\'iaz and Marathe \cite{diaz2019soft}, our network optimized the Kullback-Leibler divergence loss. The parameters of the network, the training parameters, and the smoothing function parameters were selected through random search.

We compared our model's efficiency against an adapted social force model, as previous modeling approaches have featured adaptations of the social force model \cite{ferrer2017robot,patompak_learning_2020}. Following Ferrer et al. \cite{ferrer2017robot}, we determined the model's parameters with a genetic algorithm. We used PyMoo's genetic algorithm implementation \cite{blank2020pymoo} and chose the average mean absolute error of the predictions over the training set from SocNav1 as a fitness function.
Our model, both with and without smoothing applied, outperformed the social force model (mean absolute error: social force model = 28.64, feedforward neural network = 13.90, feedforward neural network + smoothing = 13.70).
From this result, we determined that the use of the smoothing filter improved the performance of the proxemics model.

\subsection{System Overview}
\begin{figure}[!hptb]
    \centering
    \includegraphics[width=0.5\textwidth]{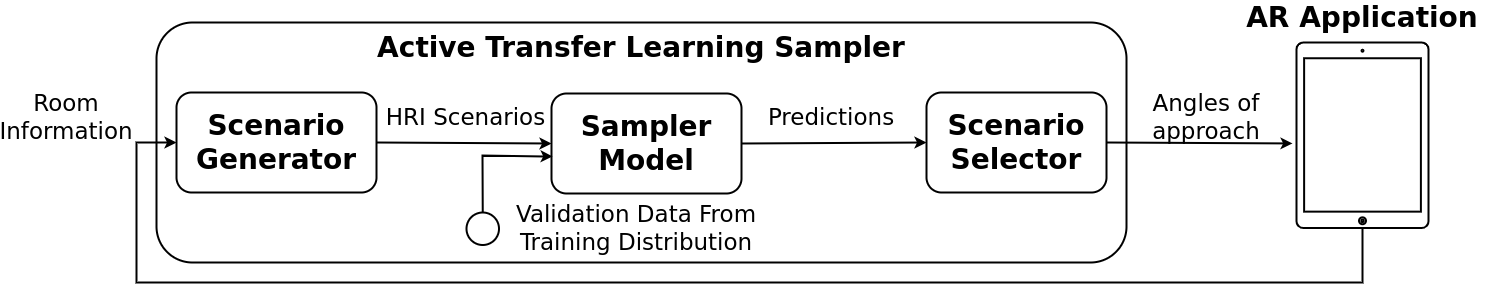}
    \caption{System Architecture Diagram}
    \label{fig:System}
\end{figure}

The implemented system comprises a tablet-based AR application and an ATL sampler. The objective of the AR application is to collect data in a way that is fast, easy, and engaging for the user. The objective of the ATL sampler is to determine which robot actions will gather the most information about the user's proxemics preferences for model personalization.

\subsubsection{AR Application}
The AR iPad application guides users through an interactive activity with a virtual projection of the Mayfield Kuri robot \cite{mayfield_robotics_explore}. Following the experimental procedure described by Patompak et al. \cite{patompak_learning_2020} and Gao et al. \cite{gao_investigating_2018}, we structured the activity to allow the virtual robot to repeatedly approach the user from different angles while the user indicates when the robot is at the most comfortable distance. The activity was designed to quickly measure the user's preferred distance from the robot while keeping them engaged and entertained throughout the interaction. We implemented a question-answering activity based on a set of virtual household objects populated throughout the virtual environment. 

 The activity proceedes through the following steps:
\begin{enumerate}
\label{activity}
    \item A virtual object appears somewhere in the AR environment;
    \item The virtual robot travels through the room to retrieve the object;
    \item The robot approaches the user with the retrieved object floating above its head;
    \item The user presses a stop sign button on the AR interface when the robot reaches the distance they feel most comfortable with, halting the robot's movement;
    \item Once stopped, the robot asks the user a question related to the object. The user selects one of two answer choices on the AR interface;
    \item The robot repeats steps 1-5 for the duration of the activity.
\end{enumerate}

A complete list of virtual objects used in the activity and the associated questions and answer options the system presented are available in Section V of the Appendix. The AR interface was installed on a 10.2" Apple iPad for the user studies described in Appendix II.

\subsubsection{ATL Sampler}

Our approach is inspired by ATL for representative sampling, in which the objective is to collect a set of labeled data points as sample data to help a model trained on a \textit{training} domain with a \textit{training} task adapt to a new \textit{application} domain and \textit{application} task.
To achieve this objective, the ATL sampler aims to select data points that are the most different from the \textit{training} domain. This is accomplished by training a model to distinguish between data points belonging to the \textit{training} domain and the \textit{application} domain. The ATL sampler leverages the model to select unlabeled data predicted with the highest confidence to belong to the \textit{application} domain.
In this context:
\begin{itemize}
\item Each data point is an HRI scenario as defined in \ref{problem_formulation};
\item Each label is the user's discomfort;
\item The \textit{Training} domain and task refer to the collection of labeled HRI scenarios taken from the SocNav1 dataset used for model training; 
\item The \textit{application} domain and task refer to the target user's preferences.
\end{itemize}
We implemented the ATL sampler to sample the robot’s angles of approach. The ATL sampler achieves this with three components: a scenario generator, a sampler model, and a scenario selector.

The sampling process proceeded as follows:
\begin{enumerate}
    \item The scenario generator receives environment information and generates a pool of HRI scenarios from the \textit{application} domain;
    \item The sampler model is trained with the generated HRI scenarios, labeled as coming from the \textit{application} domain, and validation data used in the training, labeled as coming from the \textit{training} domain;
    \item The sampler model predicts if the generated HRI scenarios come from the \textit{application} domain or the \textit{training} domain and sends the results of the predictions to the scenario selector;
    \item The scenario selector groups the generated HRI scenarios along with the model predictions by the robot's angle of approach. It then selects the instances predicted with the highest confidence to belong to the \textit{application} domain;
    \item The scenario selector communicates the chosen angles of approach to the AR interface by transmitting a list of positions, specifying the locations from which the robot should start its approach toward the participant at each angle;
    \item After the data points are labeled, the sampler model is retrained, and steps 3-5 are repeated.
\end{enumerate}

\label{AR_desc}
\section{Evaluation With User Studies}

\label{user_studies}
We conducted two concurrent studies, approved by the institution's Internal Review Board (IRB \#UP-20-00030):

\begin{itemize}
    \item Study 1: A study with a convenience population of university students to test the effectiveness of the ATL approach;
    \item Study 2: A study with the target population of older adults to verify the application’s usability and gather qualitative feedback.
\end{itemize}

\subsection{Study 1: Effectiveness Study}
The purpose of this study was to validate the ATL Sampling strategy and investigate the relationship between proxemics preferences gathered from interactions with a physical robot and those gathered through the AR activity.
During the study, participants expressed their proxemics preferences under two experimental conditions: 1) with a physical robot through verbal signals; and 2) with a virtual robot through the tablet-based AR application. The application prompted the participants to label either data points obtained through either random sampling (RS) or the ATL sampling strategy. 

\begin{figure}[!htb]
    \centering
    \begin{minipage}{0.25\textwidth}
        \centering
        \includegraphics[width=0.8\textwidth]{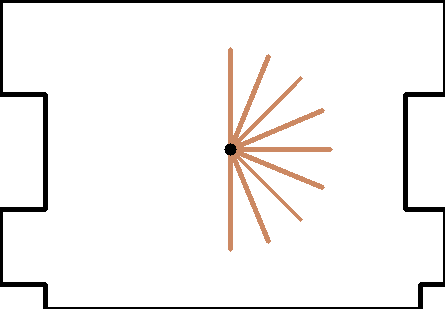}
    \end{minipage}%
    \begin{minipage}{0.25\textwidth}
        \centering
        \includegraphics[width=0.8\textwidth]{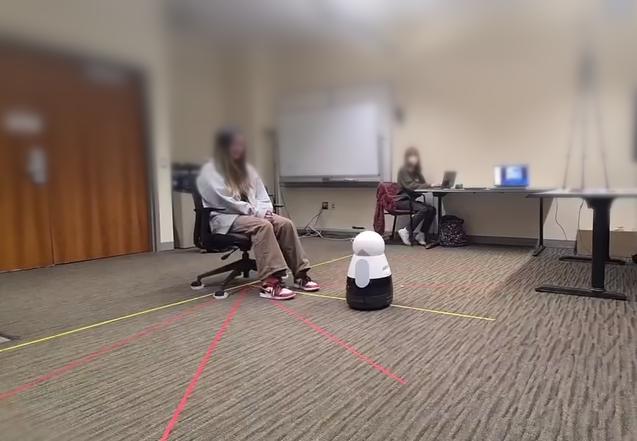}
    \end{minipage}
     \caption{Effectiveness study setup: left) a drawing of the environment and the robot approach lines, right) a view of the experiment setup}
    \label{fig:facilities_pilot}
\end{figure}

\label{study_setup}
\subsubsection{Study Setup}
The study took place in an empty university conference room, shown in Figure \ref{fig:facilities_pilot}. Participants were seated in an office chair in the center of the room. They were able to rotate the chair to look around the room, but the wheels of the chair were locked to keep their position consistent throughout the study.

\subsubsection{Participants}
25 university students aged 18-26 (M = 23.16; SD = 1.99) participated in the study. Participants self-identified their gender as 7 Women and 18 Men. They self-identified their race as 22 Asian, 1 white, and 2 did not disclose. The study took 35 minutes to complete, including the overview and setup (5 min), surveys (5 min), condition \#1 (20 min), and condition \#2 (5 min). Participants were compensated with a US \$15.00 gift card.  

\subsubsection{Study Procedure}
Participants completed a consent form and a pre-study survey with demographic and biographical questions. They were then seated in the center of the experiment room. The experiment followed a within-subject design. Participants were presented the following two conditions in a random, counter-balanced order:
\label{study_conditions}
\begin{itemize}
    \item Condition \#1: User expresses proxemics preferences for a virtual Kuri robot via the AR application administered on a 10.2" Apple iPad. This condition consisted of two phases: 1) gathering proxemics preferences for RS data points; and 2) gathering proxemics preferences for data points sampled through the ATL strategy.
    
    \item Condition \#2: User verbally expresses proxemics preferences for a physical Kuri robot. In this condition, the robot approached the user from 9 equally spaced angles between -90\textdegree{} and 90\textdegree{} with respect to the participant, indicated with 2m red lines taped on the floor (shown in Figure \ref{fig:facilities_pilot}). The Kuri robot was manually carried by a researcher to the end of the line furthest from the user, then remotely driven towards the user by a second researcher. The robot approached the participant at a rate of 0.33 m/sec. The participant was instructed to say ``stop" when the robot reached the distance they felt was most comfortable for them. The researcher then recorded the participant's chosen distance. This process was repeated for all nine angles of approach, proceeding in order from -90\textdegree{} to 90\textdegree{}.
\end{itemize}

\subsection{Study 2: Older Adult User Study}
To validate the method with the target user population of older adults, we performed a user study with fifteen older adults to gather their feedback on the AR application and their perceptions of the virtual and physical embodiments of the robot.

\subsubsection{Study Setup}
The study was set up in the same manner as the first study, described in Section \ref{study_setup} above. It was conducted in public spaces of two retirement communities in the Los Angeles area. 

\subsubsection{Participants}
We recruited fewer participants for this study due to limited time and space in the retirement communities. Fifteen residents aged 54-87 (M = 76.4; SD = 8.17) participated in the study. Participants self-identified their gender as 13 women and 2 men. They self-identified their race as 3 Black, 2 multiracial, 7 white, and 3 did not disclose. The study took 50 minutes to complete, including the overview and setup (5 min), surveys (5 min), condition \#1 (20 min), condition \#2 (5 min), SUS (10 min), and semi-structured interview (5-10 min). Participants were compensated with a US \$15.00 gift card. 

\subsubsection{Study Procedure}
Participants completed the consent form and a pre-study survey with demographic and biographical questions. They were then seated in the chair in the center of the experiment room. As in the first study, this study followed a within-subjects design. Participants were presented the same conditions described in Section \ref{study_conditions} in a random, counter-balanced order. After finishing both conditions, participants completed the System Usability Scale (SUS) \cite{brooke1996sus}. Semi-structured interviews were conducted and audio recorded. To minimize potential bias, a trained psychologist who did not participate in the study analyzed the interview transcripts. 

\section{Results}

\subsection{Study 1: Effectiveness Study Results}
The data from the effectiveness study (study 1) were evaluated in two ways. First, we compared the preferred stopping distances between the AR and physical robot conditions to validate that the virtual robot interaction can be used to approximate an interaction with the physical robot in this context. 
Second, we compared the performances of the general proxemics model with no fine-tuning, after fine-tuning with data points sampled through ATL, and after fine-tuning with RS data points. 
Results from these analyses can be found in sections \ref{physicalvsaugmented} and \ref{modelingresults}, respectively.

\subsubsection{Physical vs. AR Robot Results}
\label{physicalvsaugmented}
As a measure of similarity between preferences gathered with the virtual robot in the AR application and with the physical robot, we used the absolute difference in the mean stopping distance, following Li et al. \cite{8673116}. 

\begin{figure}[!hptb]
    \centering
    \includegraphics[width=0.5\textwidth]{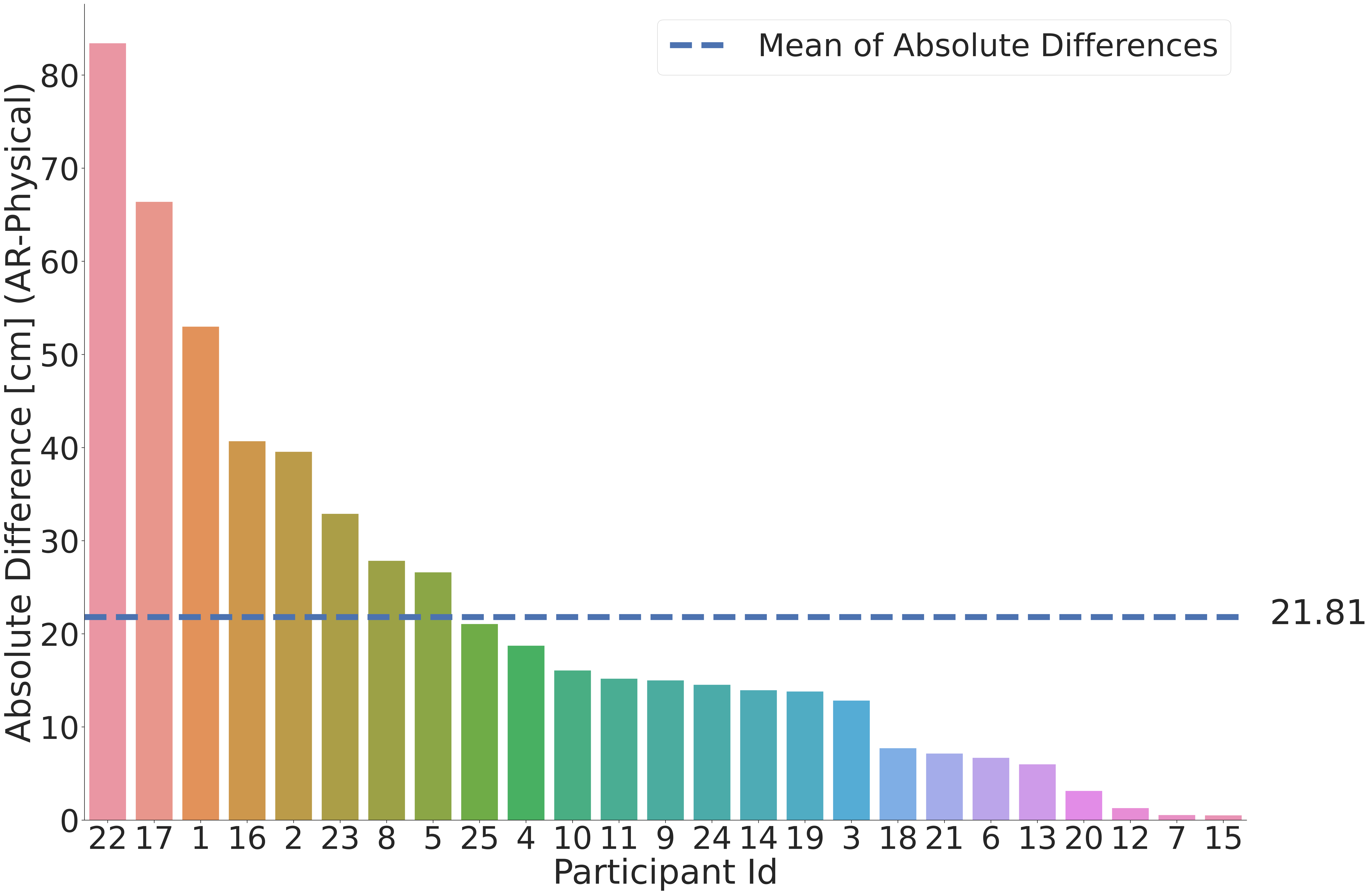}
    \caption{$\lvert Mean~stopping~distance~of~virtual~robot - mean~stopping~distance~of~physical~robot\rvert$ in the convenience population study after outliers removal}
    \label{fig:abs_diff_out}
\end{figure}

\begin{figure}[!hptb]
    \centering
    \includegraphics[width=0.5\textwidth]{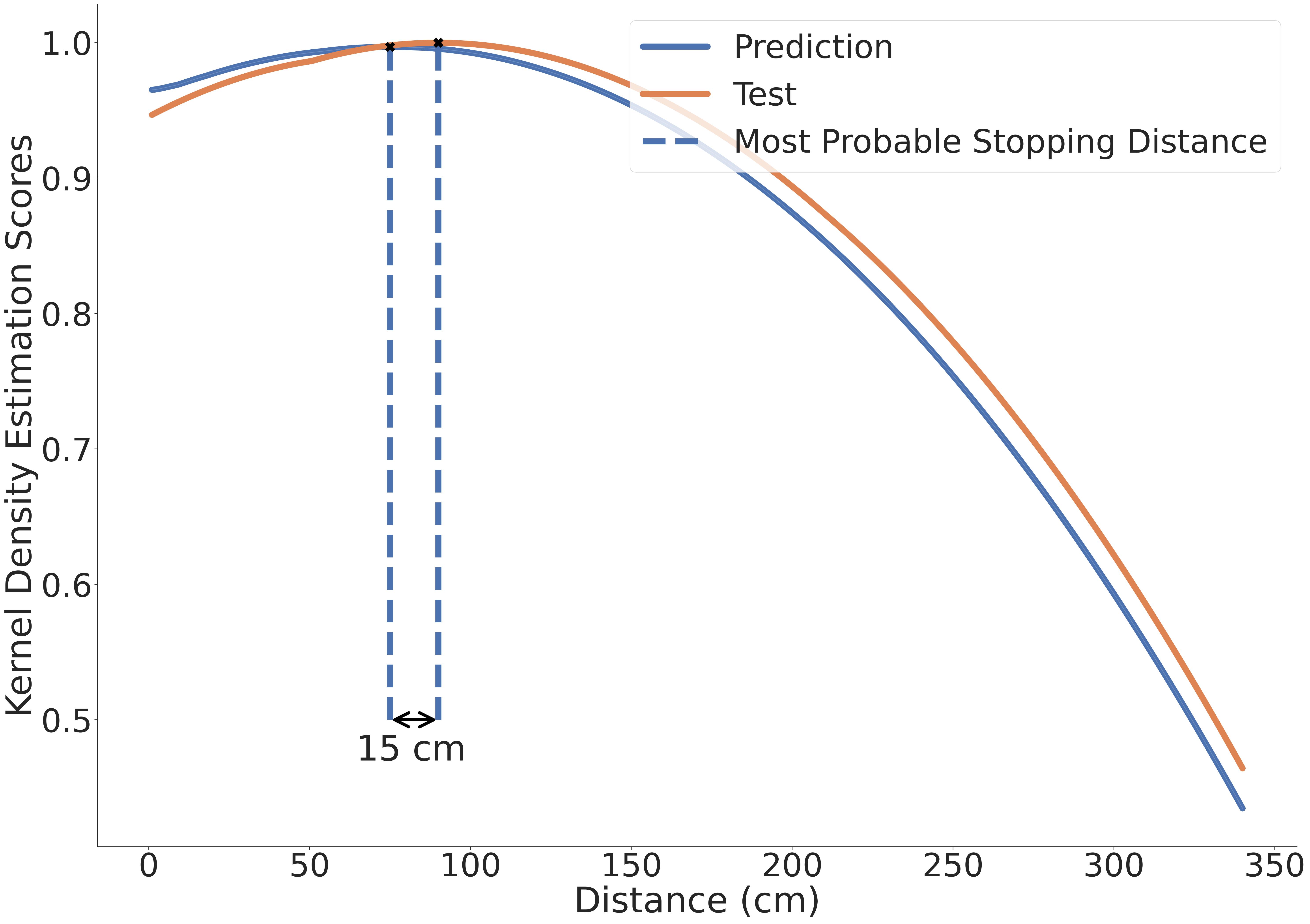}
    \caption{Example of the prediction of the density estimation for p1 in the convenience population study at an angle of $70^\circ$.}
    \label{fig:gpr_17}
\end{figure}

As a pre-processing step, we performed outlier removal using Sci-Kit Learn's implementation of Isolation Forest \cite{4781136}.
Differences between user-selected stopping distances recorded in the physical and virtual robot conditions varied greatly between participants, ranging from 0 cm to almost 100 cm. We applied kernel density estimation, following Gao et al. \cite{gao_investigating_2018}, and trained a Gaussian process regressor to characterize the relationship between the users’ stopping distances recorded with the physical robot and stopping distances recorded with the virtual robot.
We tested the regression result for the participants with a higher absolute difference (e.g., p1, p17, p22). An example of the density estimation predicted through the regression function is shown in Figure \ref{fig:gpr_17}.
We found that the difference between true and predicted stopping distances lowered for all testing participants between 58\% (p1) and 88\% (p22). By using the stopping distances of the AR robot across all participants, we improved our estimated stopping distances for the physical robot. These results suggest that the derived regression function utilizing AR stopping distances as input is a viable proxy for physical stopping distances.

\subsubsection{ATL Adaptation Results}
\label{modelingresults}

To evaluate the effectiveness of the ATL adaptation approach, we compared the performances of two different models: fine-tuned with data points gathered through ATL and fine-tuned with RS data points. 

We split the data gathered for each participant into training, validation, and test sets. The validation and test sets contained data points gathered through RS and were used to fine-tune both models. The ATL training set contained data points sampled through ATL and the RS training set contained an equal number of RS data points. Fine-tuning was performed for 50 epochs with a learning rate of 5e-5. 
During fine-tuning, we froze all layers except the output layer to preserve learned features and prevent overfitting. We applied a smoothing filter \cite{press1990savitzky} to all predictions.

\begin{figure}[!hptb]
    \centering
    \includegraphics[width=0.5\textwidth]{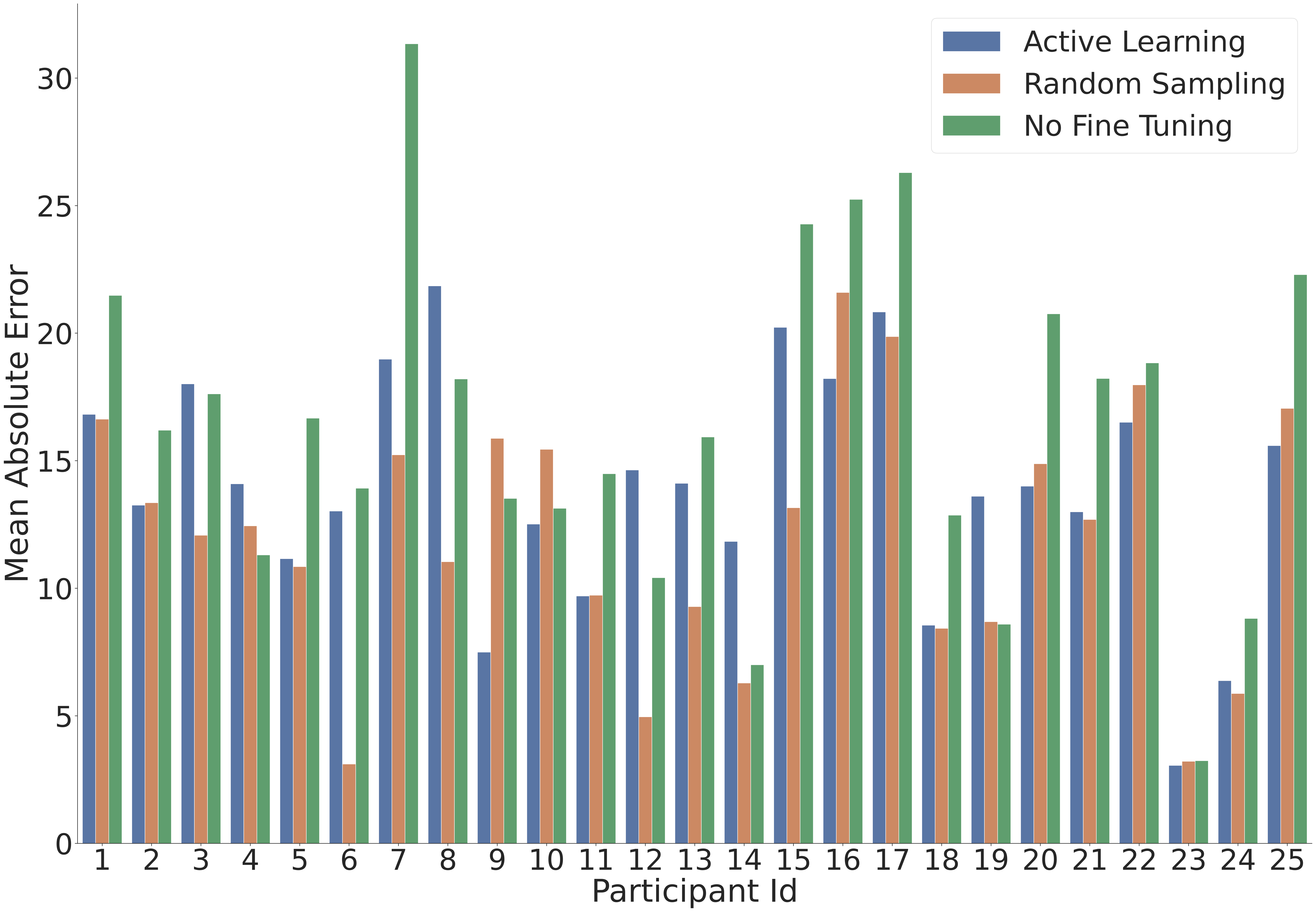}
    \caption{Mean absolute error for each participant of the convenience population study for each version of the model: fine-tuned with data points gathered through ATL, fine-tuned with data points gathered through RS, and not fine-tuned }
    \label{fig:mod_res}
\end{figure}

Figure \ref{fig:mod_res} visualizes the mean absolute error for each participant for three versions of the proxemics model: 1) fine-tuned with data points gathered through ATL; 2) fine-tuned with data points gathered through RS; and 3) not fine-tuned.

The performance of the proxemics model consistently improved after fine-tuning. However, contrary to expectation, models fine-tuned with data points obtained through RS exhibited significantly better performance than those fine-tuned with data points sampled through active learning. These observations were further supported by the results of t-tests, presented in Table \ref{table:model_error}.

\begin{table}[!t]
\begin{center}
\caption{Results of lower tailed paired t-tests between Mean absolute error for each participant of the convenience population study for each version of the model: fine-tuned with data points gathered through ATL, fine-tuned with data points gathered through RS, and not fine-tuned ( $\neg$FT)}
\label{table:model_error}
\begin{tabular}{| c | c | c | c | c | c | c |}
\hline
A & B & $\mu$/$\sigma$ A & $\mu$/$\sigma$ B & p-val & df & t \\
\hline
RS & ATL & 11.99(5.00) & 13.90(4.60) & .02 & 23 & -2.10\\
RS & $\neg$FT & 11.99(5.00) & 16.43(6.58) & $<$.001 & 23 & -5.04 \\
ATL & $\neg$FT & 13.90(4.60) & 16.43(6.58) & .004 & 23 & -2.87 \\
ATL & RS & 13.90(4.60) & 11.99(5.00) & .98 & 23 & 2.10\\
\hline
\end{tabular}
\end{center}
\end{table}

After reexamining the study data to understand the decreased performance of the ATL fine-tuned model, we formed a {\it post hoc} hypothesis that models trained on data points sampled at a higher variety of angles would yield better-performing fine-tuned models. To test this hypothesis, we computed Spearman’s rank correlation to assess the relationship between the standard deviation of the angles of the training data points and their respective models' mean absolute error for each participant. 
There was a negative correlation between the two variables, $r(23)=-0.30, p=0.04$. These values indicate a weak monotonic relationship between the standard deviation in angles of sampled data points and the mean absolute error of the model; therefore, our first {\it post hoc} hypothesis was supported.
 
\begin{figure}[!hptb]
    \centering
    \includegraphics[width=0.5\textwidth]{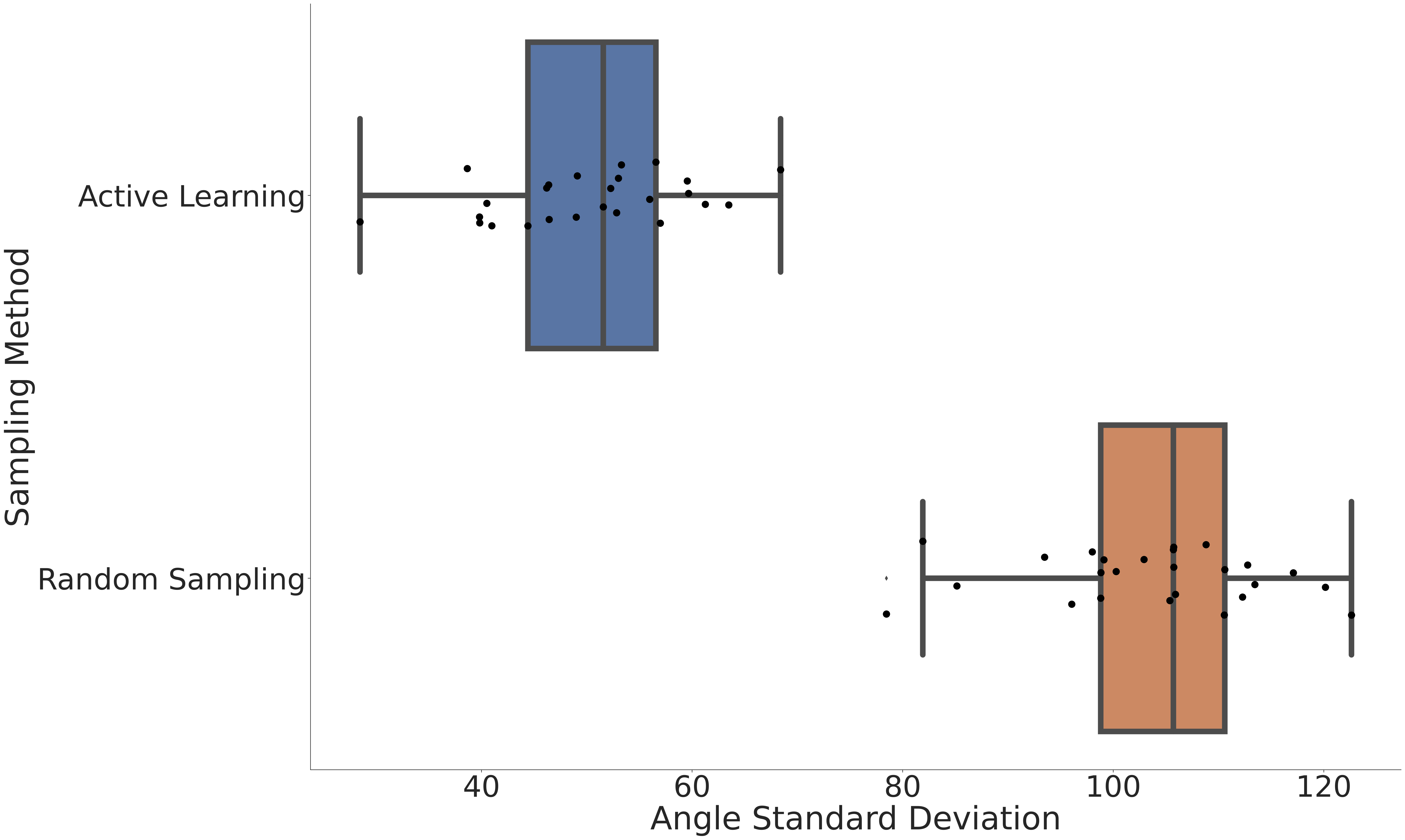}
    \caption{Angle standard deviation for the active learning and RS methods}
    \label{fig:std_angle}
\end{figure}

As shown in Figure \ref{fig:std_angle}, RS data points consistently showed a higher standard deviation in sampling angle compared to data points sampled through ATL.
This discrepancy may have arisen because RS uniformly selects data points from across the space, while the ATL approach prioritizes sampling from regions where data points differ the most from those encountered during training, potentially limiting the diversity of acquired points.

We also formed a second {\it post hoc} hypothesis that the poorer performance of the model fine-tuned with ATL-sampled data points was related to the disparities in the distributions of data points obtained through ATL and RS. 
Given that disparities in distributions primarily stem from variations in the ranges of angles sampled in the space, we hypothesized that if ATL sampled significantly fewer angles than RS for the same participant, the results were inferior compared to other participants with less disparity in sampled angles.
To test that hypothesis, we employed a Kolmogorov test and utilized the test statistics as a measure of similarity between the distributions. Subsequently, we performed a Spearman correlation between the statistics from the Kolmogorov test and the performances of the models trained with the data points obtained through ATL for each participant. Our findings indicate a moderate monotonic relationship between these two variables $r(23)=0.54,  p=0.005$; therefore our second {\it post hoc} hypothesis is supported.

\subsection{Study 2: Older Adult User Study Results}

The data from the older users study (study 2) were evaluated using the SUS \cite{brooke1996sus} and qualitative analysis of post-study interviews. 

The SUS generates a usability score from 0 to 100 that is directly proportional to the perceived system usability, with higher scores indicating greater perception of usability. We administered the SUS with all 15 study participants and found a mean score of 66\%.

Semi-structured post-study interviews with the older adult participants yielded several insights about their perceptions of virtual and physical robots. 
Overall, the participants positively evaluated the AR interface. 
Participants appreciated the tablet's touchscreen (n=1), and found the application design ``cute", ``fun", and ``pleasurable" (n=5). They described the UI as ``consistent" with ``good navigation," and the participants felt that they ``knew what [they were] doing while using it" (n=3). They also reported that learning to use the application improved their confidence adopting new technology, and that the activity kept their mind active (n=3). 

Six participants suggested ways to improve the system, pointing out that the tablet was ``heavy" (n=2) and that the touchscreen was too ``sensitive" (n=1). Some reported difficulties getting used to using the AR interface. They initially felt they were ``chasing the robot around" (n=1), but eventually grew more comfortable: ``it took me a while to figure it out but once I did it was fine" (n=1).

46.7\% (n=7) of participants reported that understanding how the robot learned their preferences from the tablet interface improved their trust in the robot, while 33.33\% (n=5) pointed out that the robot was already perceived as trustworthy. P3 said, ``[the question] suggests that I didn't trust the robot in the first place, and I just thought it was a really cute little thing. Absolutely trustworthy." 
60\% (n=9) said they would enthusiastically use the tablet interface to personalize other aspects of the robot’s social behavior. 
One participant was deeply skeptical about the extensive use of robots, stating that they prefer ``human communication."  

\section{Discussion}

In the ATL effectiveness study, we 1) investigated the difference in preferred stopping distances between the AR and physical robot conditions; and 2) evaluated the performances of the proxemics model with no fine-tuning, after fine-tuning with data points sampled through ATL, and after fine-tuning with data points sampled through RS.

We found a relationship between the data points gathered through the physical robot and AR robot, improving our stopping distance estimations for the study participants between 58\% and 88\% percent.
Our study was designed to investigate the use of ATL for personalized proxemics models, with a secondary goal of  exploring of the relationship between proxemics preferences for virtual and physical robots. Future work should seek to further investigate this relationship. 

Our results support findings from prior approaches that personalized proxemics models led to improved performance. Fine-tuning the proxemics models with individual user preferences lowered the error on the test set by 26.97\% on average across all participants.

A key question emerged from the results:
{\it Why did the RS approach perform significantly better than active learning?} To answer this question we investigated: 1) the connection between the variability in sampling angles and model performance; and 2) the differences in distributions between the data points gathered through ATL and RS. First, we found that models trained on data points sampled from a higher variety of angles would yield better-performing fine-tuned models. Thus, models trained on RS data points would yield higher performances since the RS data points exhibited a higher standard deviation in the sampling angles with respect to the data points sampled through active learning (see Figure \ref{fig:std_angle}). 
Second, we found that models trained using data points acquired through ATL exhibited worse performance as the differences between the ATL and RS distributions increased. 

From the user study with older adults, we found that the AR application was well received. The mean SUS score of 66 is slightly below the average reported SUS score of 68 \cite{brooke1996sus}, and the post-study interviews indicated that the study participants enjoyed the interaction with both the physical and the virtual robot, validating our personalized proxemics modeling approach. 46.7\% of participants indicated that our approach helped them to build trust with the physical robot, and 60\% of participants expressed interest in using the AR interface to personalize more robot behaviors. This feedback can inform the design of AR interfaces for older adult users, including selecting a lighter tablet, and allowing the participants to familiarize themselves with the interface prior to the data collection period.

\subsection{Limitations and Future Research Directions}

It is possible that shortcomings in the methods of collecting proxemics preferences in both the virtual and physical robot conditions were reflected in the collected data. It is likely that there was a difference between the ground truth preference of the user and the recorded preference, in both the virtual and physical robot conditions. In the physical robot condition, there was a ~1-second delay between the time that the user performed the stop gesture and the time that the approaching robot halted. In the virtual robot condition, because the virtual robot was halted by a user touch event in the AR interface, the delay was negligible. However, it is possible that in both conditions, a delay occurred due to the user's reaction time (e.g., a delay between the time at which the robot arrived at the most comfortable position for the user and the time at which the user signaled to the robot to stop). It is also possible that participants began to anticipate this delay and account for it by performing the stop gesture prematurely. In our analysis, we made the assumption that any such discrepancies from users learning the system would be removed with outlier removal; however, future work should validate this further.

In general, our work sought to perform a preliminary exploration of the relationship between user proxemics preferences with a virtual robot and their preferences with a physical robot; future work should seek to explore this relationship further. Decreasing the difference between preferences recorded with virtual and physical robots would likely help produce models that more readily generalize from virtual to physical robot interactions.
While the approach we presented is flexible, our user studies did not evaluate the approach with different robot embodiments beyond the Kuri robot. Future work may include validating the system's robustness with different robot embodiments, as well as further improving the interaction to foster a more open, rapport-building dialogue between the user and robot.

 \section{Conclusion}
This paper presented a method for collecting users' proxemics preferences for mobile robots through an engaging AR interface. Our approach is tailored for older adult users, and aims to encourage their use of socially assistive robots. Toward that end, we presented and evaluated an ATL approach to form personalized deep learning proxemics models that reflect preferences of individual users. Our method addresses gaps in the state of the art by enabling personalization from user feedback while considering contextual cues, such as position in the environment and room shape and size.

We evaluated the AR interface and the personalized modeling approach in two user studies, one with participants from a convenience population and one with older adult participants.  Specifically, we assessed the effectiveness of the ATL approach with a convenience population. We evaluated the system's usability and gathered qualitative feedback from the target population of older adults. Results from the convenience population study demonstrated a significant improvement in model performance following fine-tuning, although we found no significant difference in performance between the ATL and RS strategies.

The results from the user study with older adults showed that they reacted positively to the activity and AR interface. Participant feedback suggested that they appreciated the system's usability, enjoyed the interaction with both the physical and the virtual robot, and were interested in using the AR interface to personalize other aspects of the robot's behavior. Overall, the described method overcomes several current shortcomings of the state of the art and provides directions for future research and development of user-centric methods for HRI. 

\section{Acknowledgments}
This work was supported by the National Science Foundation National Robotics Initiative 2.0 grant for ``Communicate, Share, Adapt: A Mixed Reality Framework for Facilitating Robot Integration and Customization'', NSF IIS-1925083. Anna-Maria Velentza was funded by the Fulbright Foundation, Greece.

 
\bibliographystyle{IEEEtran}
\bibliography{bibliograph}

\begin{thebibliography}{10}
\providecommand{\url}[1]{#1}
\csname url@samestyle\endcsname
\providecommand{\newblock}{\relax}
\providecommand{\bibinfo}[2]{#2}
\providecommand{\BIBentrySTDinterwordspacing}{\spaceskip=0pt\relax}
\providecommand{\BIBentryALTinterwordstretchfactor}{4}
\providecommand{\BIBentryALTinterwordspacing}{\spaceskip=\fontdimen2\font plus
\BIBentryALTinterwordstretchfactor\fontdimen3\font minus \fontdimen4\font\relax}
\providecommand{\BIBforeignlanguage}[2]{{%
\expandafter\ifx\csname l@#1\endcsname\relax
\typeout{** WARNING: IEEEtran.bst: No hyphenation pattern has been}%
\typeout{** loaded for the language `#1'. Using the pattern for}%
\typeout{** the default language instead.}%
\else
\language=\csname l@#1\endcsname
\fi
#2}}
\providecommand{\BIBdecl}{\relax}
\BIBdecl

\bibitem{national2020social}
{National Academies of Sciences, Engineering, and Medicine}, \emph{Social isolation and loneliness in older adults: Opportunities for the health care system}.\hskip 1em plus 0.5em minus 0.4em\relax National Academies Press, 2020.

\bibitem{pino2015we}
M.~Pino, M.~Boulay, F.~Jouen, and A.-S. Rigaud, ``“are we ready for robots that care for us?” attitudes and opinions of older adults toward socially assistive robots,'' \emph{Frontiers in aging neuroscience}, vol.~7, p. 141, 2015.

\bibitem{okita2012captain}
S.~Y. Okita, V.~Ng-Thow-Hing, and R.~K. Sarvadevabhatla, ``Captain may i? proxemics study examining factors that influence distance between humanoid robots, children, and adults, during human-robot interaction,'' in \emph{Proceedings of the seventh annual ACM/IEEE international conference on Human-Robot Interaction}, 2012, pp. 203--204.

\bibitem{eresha2013investigating}
G.~Eresha, M.~H{\"a}ring, B.~Endrass, E.~Andr{\'e}, and M.~Obaid, ``Investigating the influence of culture on proxemic behaviors for humanoid robots,'' in \emph{2013 IEEE Ro-Man}.\hskip 1em plus 0.5em minus 0.4em\relax IEEE, 2013, pp. 430--435.

\bibitem{bhagya2019exploratory}
S.~Bhagya, P.~Samarakoon, M.~Viraj, J.~Muthugala, A.~Buddhika, P.~Jayasekara, and M.~R. Elara, ``An exploratory study on proxemics preferences of humans in accordance with attributes of service robots,'' in \emph{2019 28th IEEE International Conference on Robot and Human Interactive Communication (RO-MAN)}.\hskip 1em plus 0.5em minus 0.4em\relax IEEE, 2019, pp. 1--7.

\bibitem{yasumoto2011personal}
M.~Yasumoto, H.~Kamide, Y.~Mae, K.~Kawabe, S.~Sigemi, M.~Hirose, and T.~Arai, ``Personal space of humans in relation with humanoid robots depending on the presentation method,'' in \emph{2011 IEEE/SICE International Symposium on System Integration (SII)}.\hskip 1em plus 0.5em minus 0.4em\relax IEEE, 2011, pp. 797--801.

\bibitem{walters2007robot}
M.~L. Walters, K.~L. Koay, S.~N. Woods, D.~S. Syrdal, and K.~Dautenhahn, ``Robot to human approaches: Preliminary results on comfortable distances and preferences.'' in \emph{AAAI Spring Symposium: Multidisciplinary Collaboration for Socially Assistive Robotics}, 2007, p. 103.

\bibitem{gao_investigating_2018}
Y.~Gao, S.~Wallkötter, M.~Obaid, and G.~Castellano, ``Investigating {Deep} {Learning} {Approaches} for {Human}-{Robot} {Proxemics},'' in \emph{2018 27th {IEEE} {International} {Symposium} on {Robot} and {Human} {Interactive} {Communication} ({RO}-{MAN})}, Aug. 2018, pp. 1093--1098, iSSN: 1944-9437.

\bibitem{patompak_learning_2020}
\BIBentryALTinterwordspacing
P.~Patompak, S.~Jeong, I.~Nilkhamhang, and N.~Y. Chong, ``\BIBforeignlanguage{en}{Learning {Proxemics} for {Personalized} {Human}–{Robot} {Social} {Interaction}},'' \emph{\BIBforeignlanguage{en}{International Journal of Social Robotics}}, vol.~12, no.~1, pp. 267--280, Jan. 2020. [Online]. Available: \url{https://doi.org/10.1007/s12369-019-00560-9}
\BIBentrySTDinterwordspacing

\bibitem{monarch2021human}
R.~M. Monarch, \emph{Human-in-the-Loop Machine Learning: Active learning and annotation for human-centered AI}.\hskip 1em plus 0.5em minus 0.4em\relax Simon and Schuster, 2021.

\bibitem{zhuang2020comprehensive}
F.~Zhuang, Z.~Qi, K.~Duan, D.~Xi, Y.~Zhu, H.~Zhu, H.~Xiong, and Q.~He, ``A comprehensive survey on transfer learning,'' \emph{Proceedings of the IEEE}, vol. 109, no.~1, pp. 43--76, 2020.

\bibitem{manso2020socnav1}
L.~J. Manso, P.~Nu{\~n}ez, L.~V. Calderita, D.~R. Faria, and P.~Bachiller, ``Socnav1: A dataset to benchmark and learn social navigation conventions,'' \emph{Data}, vol.~5, no.~1, p.~7, 2020.

\bibitem{press1990savitzky}
W.~H. Press and S.~A. Teukolsky, ``Savitzky-golay smoothing filters,'' \emph{Computers in Physics}, vol.~4, no.~6, pp. 669--672, 1990.

\bibitem{diaz2019soft}
R.~Diaz and A.~Marathe, ``Soft labels for ordinal regression,'' in \emph{Proceedings of the IEEE/CVF conference on computer vision and pattern recognition}, 2019, pp. 4738--4747.

\bibitem{ferrer2017robot}
G.~Ferrer, A.~G. Zulueta, F.~H. Cotarelo, and A.~Sanfeliu, ``Robot social-aware navigation framework to accompany people walking side-by-side,'' \emph{Autonomous Robots}, vol.~41, no.~4, pp. 775--793, 2017.

\bibitem{blank2020pymoo}
J.~Blank and K.~Deb, ``Pymoo: Multi-objective optimization in python,'' \emph{IEEE Access}, vol.~8, pp. 89\,497--89\,509, 2020.

\bibitem{mayfield_robotics_explore}
\BIBentryALTinterwordspacing
``\BIBforeignlanguage{en-US}{Explore the {Technology} {Behind} {Kuri}, {The} {Home} {Robot}}.'' [Online]. Available: \url{https://www.heykuri.com/explore-kuri/}
\BIBentrySTDinterwordspacing

\bibitem{brooke1996sus}
J.~Brooke \emph{et~al.}, ``Sus-a quick and dirty usability scale,'' \emph{Usability Evaluation in Industry}, vol. 189, no. 194, pp. 4--7, 1996.

\bibitem{8673116}
R.~Li, M.~van Almkerk, S.~van Waveren, E.~Carter, and I.~Leite, ``Comparing human-robot proxemics between virtual reality and the real world,'' in \emph{2019 14th ACM/IEEE International Conference on Human-Robot Interaction (HRI)}, 2019, pp. 431--439.

\bibitem{4781136}
F.~T. Liu, K.~M. Ting, and Z.-H. Zhou, ``Isolation forest,'' in \emph{2008 Eighth IEEE International Conference on Data Mining}, 2008, pp. 413--422.

\end{thebibliography}

\begin{IEEEbiographynophoto}{Maja Matari\'c}
Maja Matari\'c is the Chan Soon-Shiong Distinguished Professor of Computer Science, Neuroscience, and Pediatrics at the University of Southern California (USC), and founding director of the USC Robotics and Autonomous Systems Center. Her Ph.D. and MS are from MIT, BS from Kansas University. She is Fellow of AAAS, IEEE, AAAI, and ACM, member of the American Academy of the Arts and Sciences, recipient of the Presidential Award for Excellence in Science, Mathematics \& Engineering Mentoring, Anita Borg Institute Women of Vision for Innovation, NSF Career, MIT TR35 Innovation, and IEEE RAS Early Career Awards, and authored ``The Robotics Primer" (MIT Press). She led the USC Viterbi K-12 STEM Center and actively mentors K-12 students, women, and other underrepresented groups toward pursuing STEM careers. Her university leadership includes serving as vice president of research (2020-21) and as engineering vice dean of research (2006-19).  A pioneer of the field of socially assistive robotics, her research is developing human-machine interaction methods for personalized support in convalescence, rehabilitation, training, and education for autism spectrum disorders, stroke, dementia, anxiety, and other major health and wellness challenges.  
\end{IEEEbiographynophoto}

\begin{IEEEbiographynophoto}{Massimiliano Nigro}
Massimiliano is a Ph.D. student of Computer Science and Engineering at Politecnico di Milano. He was a visiting researcher in the USC Interaction lab in 2022. He is interested in exploring how to leverage AR and machine learning to develop personalized models for robot behavior. 
\end{IEEEbiographynophoto}

\begin{IEEEbiographynophoto}{Amy O'Connell}
Amy O'Connell is a third year Ph.D. student in the Computer Science department at the University of Southern California. Before joining the Interaction Lab at USC, she received her BA in Computer Science from Vassar College.
\end{IEEEbiographynophoto}

\begin{IEEEbiographynophoto}{Thomas Groechel}
Thomas Groechel was a Ph.D. student in the Interaction Lab from 2018 to 2023. Tom is now working as a fulltime software engineer in industry.
\end{IEEEbiographynophoto}

\begin{IEEEbiographynophoto}{Anna-Maria Velentza}
Anna-Maria is a Fulbright scholar and Ph.D. candidate from the Laboratory of Informatics and Robotics for Education and Society (LIRES) at the University of Macedonia. She is a Psychologist with an M.Sc. in Computational Neuroscience and Cognitive Robotics from the University of Birmingham. Her research mainly focuses on human-robot interaction and the role of socially assistive robots in human cognitive functions such as memory and attention, social applications, and how to keep humans in the loop in robotics and  Cyber Physical Systems design process.
\end{IEEEbiographynophoto}
 
\vspace{11pt}




\vfill

\end{document}


\onecolumn
\title{Appendices}
\maketitle
\section{Data Pre-Processing}
\begin{figure}[!hptb]
    \centering
    \includegraphics[width=0.5\textwidth]{Nigro 2024 An Interactive Augmented Reality Interface for Personalized Proxemics 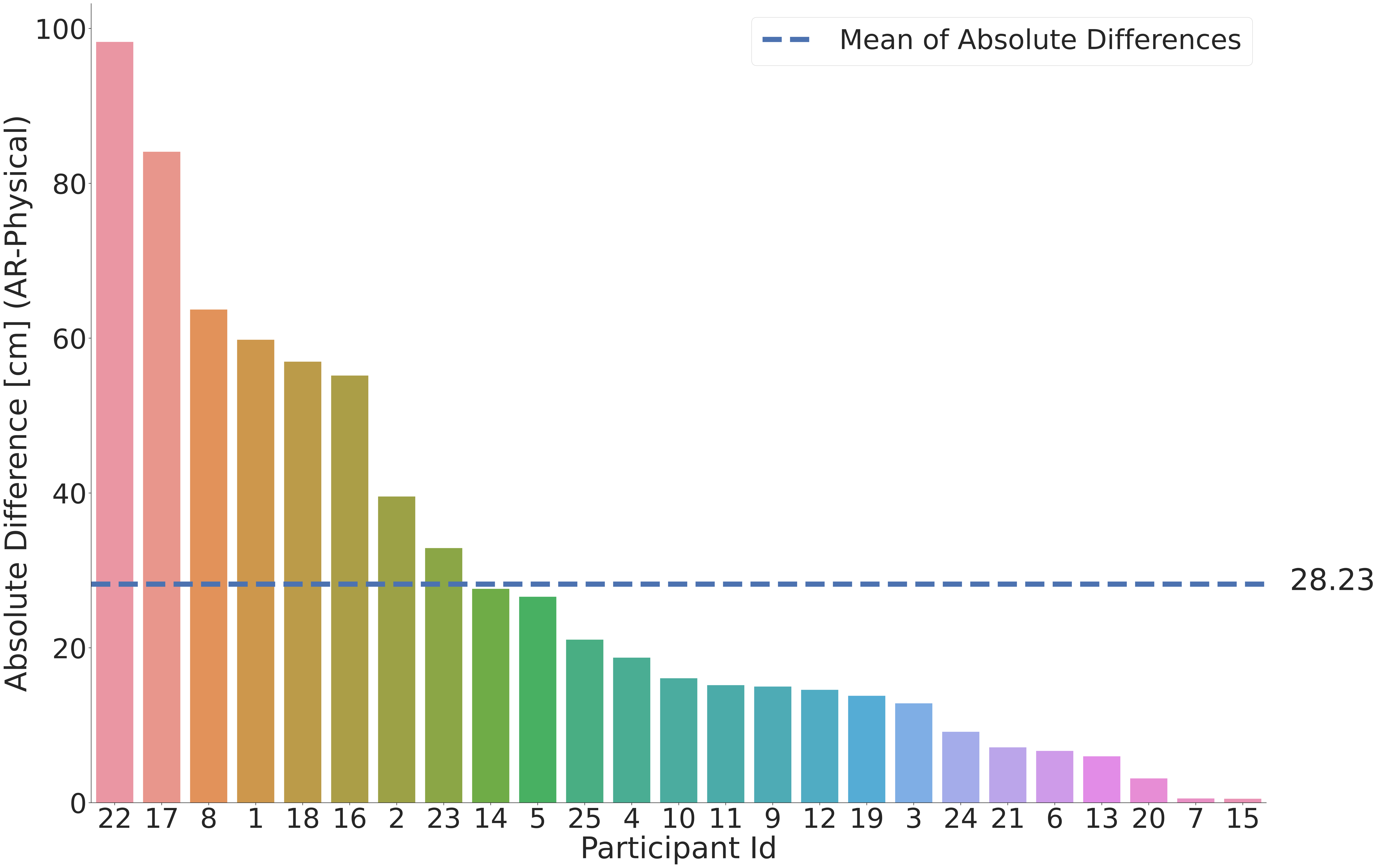}
    \caption{$\lvert Mean~stopping~distance~of~virtual~robot - mean~stopping~distance~of~physical~robot\rvert$ in the convenience population study}
    \label{fig:abs_diff}
\end{figure}

As a measure of similarity between preferences gathered with the virtual robot in the AR application and the physical robot, we used the absolute difference in the mean stopping distance.
Some participants shared very similar stopping distance preferences, with close to the same means (e.g., participants 7 and 15), while others had very different preferences, with differences of distances as large as 100 cm (see Figure \ref{fig:abs_diff}).

The observed difference in the means could be due to outliers present in the data. We therefore performed outlier removal, applying Sci-Kit Learn Isolation Forest.
\begin{figure}[!hptb]
    \centering
    \includegraphics[width=0.5\textwidth]{Nigro 2024 An Interactive Augmented Reality Interface for Personalized Proxemics Modeling/images/results/improved_difference.png}
    \caption{$\lvert Mean~stopping~distance~of~virtual~robot - mean~stopping~distance~of~physical~robot\rvert$ in the convenience population study after outliers removal}
    \label{fig:abs_diff_out}
\end{figure}
After removing outliers, the average absolute difference in the mean stopping distance across all participants difference decreased by 23\%, though the impact of the outlier removal was different among participants. For example, the differences in mean stopping distance for participants 12 and 18 decreased by more than the 87.5\%, while the differences in mean stopping distance for participants 1, 17 and 22 decreased by only 15\% to 25\% (see Figure \ref{fig:abs_diff_out}).
\section{Kernel Density Estimation and Gaussian Process Regressor}
We chose the Sci-Kit Learn's implementation for the kernel density estimation and used cross-validation to determine the kernel type and parameters. We used the stopping distances gathered through the physical robot and chose as a metric the mean of the differences between the distance from the person of the point corresponding with the maximum density estimated (estimated stopping distance) and the mean of the preferred stopping distances. Thus, we chose a Gaussian kernel with a bandwidth of 1. We used Sci-Kit Learn's implementation of a Gaussian process regressor and selected the kernel through cross-validation using as metric the difference in stopping distances for the physical robot estimated through test and predicted density estimates. We chose a Rational Quadratic kernel with parameters alpha and length equal to 1.
\section{Proxemics Model Parameters}
\begin{itemize}
    \item \textbf{Network}: 3 hidden layers, 64 neurons, 1e-1 weight decay;
    \item \textbf{Training parameters}: 0.6 momentum, 0.5e-3 learning rate;
    \item \textbf{Smoothing function}: polyorder of 1, window size 61.
\end{itemize}

\section{Stopping Distances Variation}
\begin{table}[htbp]
    \centering
    \begin{tabular}{|l|r|r|}
        \toprule
        Participant Id & Variation Before Outlier Removal &  Variation After Outlier Removal \\
        \midrule
        1 &                 0.40 &                0.42 \\
        \hline
        2 &                 0.34 &                0.34 \\
        \hline
        3 &                 0.23 &                0.23 \\
        \hline
        4 &                 0.23 &                0.23 \\
        \hline
        5 &                 0.26 &                0.26 \\
        \hline
        6 &                 0.34 &                0.34 \\
        \hline
        7 &                 0.17 &                0.17 \\
        \hline
        8 &                 0.62 &                0.57 \\
        \hline
        9 &                 0.24 &                0.24 \\
        \hline
        10 &                 0.25 &                0.25 \\
        \hline
        11 &                 0.20 &                0.20 \\
        \hline
        12 &                 0.45 &                0.34 \\
        \hline
        13 &                 0.24 &                0.24 \\
        \hline
        14 &                 0.51 &                0.42 \\
        \hline
        15 &                 0.30 &                0.30 \\
        \hline
        16 &                 0.53 &                0.55 \\
        \hline
        17 &                 0.43 &                0.38 \\
        \hline
        18 &                 0.83 &                0.88 \\
        \hline
        19 &                 0.28 &                0.28 \\
        \hline
        20 &                 0.21 &                0.21 \\
        \hline
        21 &                 0.21 &                0.21 \\
        \hline
        22 &                 0.46 &                0.48 \\
        \hline
        23 &                 0.31 &                0.31 \\
        \hline
        24 &                 0.35 &                0.24 \\
        \hline
        25 &                 0.24 &                0.24 \\
        \bottomrule
    \end{tabular}
    \caption{Variation before and after outlier removal of AR robot stopping distances for all participants}
    \label{tab:my_table}
\end{table}

\section{Interaction Dialogue}
\begin{table}[htbp]
    \centering
    \caption{The following virtual objects and corresponding dialogue appeared in the AR interaction:}
    \begin{tabular}{|l|p{3cm}|p{2cm}|p{7cm}|}
        \hline
        \textbf{Object} & \textbf{Question} & \textbf{Answer Choices} & \textbf{Response} \\
        \hline
        Box T & Do you want to know what’s in this box? & Yes \newline No & I can’t tell you. It’s a secret. \newline Ok, I won’t tell you. \\
        \hline
        Shoe T & Do you like to wear shoes? & Yes \newline No & Wow, I wish I could wear shoes. \newline Me neither! I don’t even have feet! \\
        \hline
        Chess Board T & Do you know how to play chess? & Yes \newline No & Me too! We should play some time. \newline We should play sometime so I can teach you! \\
        \hline
        Ping pong paddle & Do you like to play ping pong? & Yes \newline No & I bet you’re really good at it. \newline Me neither! I always hit the ball too hard. \\
        \hline
        Badminton racket & Do you like to play badminton? & Yes \newline No & I bet it would be very fun to play badminton with you. \newline Me neither. I can never get the birdie over the net. \\
        \hline
        Broom & Do you like to sweep? & Yes \newline No & I bet your floor is very clean. \newline Me neither. I prefer to vacuum. \\
        \hline
        Guitar & Can you play the guitar? & Yes \newline No & I would love to hear you play. \newline Me neither, but I really love the sound! \\
        \hline
        Scissors & Do you own a pair of scissors? & Yes \newline No & I bet you can cut lots of stuff with them! \newline You can borrow mine if you need some. \\
        \hline
        Pencil & Do you prefer to write with a pencil or a pen? & Pencil \newline Pen & Me too! I like to erase my mistakes. \newline I think there is a pen around here somewhere. \\
        \hline
        Pen & Do you prefer to write in cursive or print? & Cursive \newline Print & Me too! I like my letters to look fancy. \newline I think print is easier to read. \\
        \hline
        Calculator & Are you good at math? & Yes \newline No & Me too! All robots are good at math. \newline That’s ok. That's what calculators are for! \\
        \hline
        Notepad & Do you use a notebook often? & Yes \newline No & Me too! I write things down so I won’t forget them. \newline You should try it. It is fun to write and draw. \\
        \hline
        Stapler & Do you prefer to tape or staple papers together? & Tape \newline Staple & Tape can hold a lot of things together! \newline Me too! Tape always sticks to me. \\
        \hline
        Box T & Do you want to know what’s in this box? & Yes \newline No & I can’t tell you. It’s a secret. \newline Ok, I won’t tell you. \\
        \hline
        Shoe T & Do you like to wear shoes? & Yes \newline No & Wow, I wish I could wear shoes. \newline Me neither! I don’t even have feet! \\
        \hline
        Chess Board T & Do you know how to play chess? & Yes \newline No & Me too! We should play some time. \newline We should play sometime so I can teach you! \\
        \hline
        Ping pong paddle & Do you like to play ping pong? & Yes \newline No & I bet you’re really good at it. \newline Me neither! I always hit the ball too hard. \\
        \hline
        Badminton racket & Do you like to play badminton? & Yes \newline No & I bet it would be very fun to play badminton with you. \newline Me neither. I can never get the birdie over the net. \\
        \hline
        Broom & Do you like to sweep? & Yes \newline No & I bet your floor is very clean. \newline Me neither. I prefer to vacuum. \\
        \hline
        Guitar & Can you play the guitar? & Yes \newline No & I would love to hear you play. \\
        \hline
        Apple & Do you like red or green apples? & Red \newline Green & Me too! I like honey crisp. \newline I wish there was a green apple around here for you! \\
        \hline
        Plate & Have you used a plate today? & Yes \newline No & I hope your meal was very tasty. \newline Me neither. I always eat out of bowls. \\
        \hline
        Fork & If you could only eat with forks or spoons for the rest of your life, which would you choose? & Forks \newline Spoons & Eating soup would be very difficult. \newline Eating spaghetti would be very difficult. \\
        \hline
        Spoon & Do you prefer to eat with big or little spoons? & Big \newline Little & Big spoons can hold more food. \newline Little spoons are easier to carry. \\
        \hline
        Cleaver & Do you like to cook? & Yes \newline No & I bet your food is very yummy! \newline Me neither. I prefer baking! \\
        \hline
        Rolling Pin & Do you like to bake? & Yes \newline No & Me too! My favorite thing to bake is cookies. \newline That’s ok. I can bake some things for us to share. \\
        \hline
        Ladle & Do you prefer tomato soup or chicken noodle soup? & Tomato \newline Chicken noodle & Me too! I like to eat mine with a grilled cheese sandwich. \newline Chicken noodle soup always makes me feel better. \\
        \hline
        Watermelon & Do you prefer big or small watermelons? & Big \newline Small & Me too! I love sharing watermelon with friends. \newline Big watermelons can be too heavy to pick up. \\
        \hline
        Glass & Do you prefer to drink juice or water? & Juice \newline Water & Me too! I drink orange juice every morning. \newline Water is very refreshing on a hot day. \\
        \hline
        Milk & Do you like milk in your coffee? & Yes \newline No & Black coffee is too bitter for me. \newline Me neither. All robots are lactose intolerant. \\
        \hline
            \end{tabular}
        \end{table}

\begin{table}[htbp]
    \centering
    \caption{The following virtual objects and corresponding dialogue appeared in the AR interaction:}
    \begin{tabular}{|l|p{3cm}|p{2cm}|p{7cm}|}
        \hline
        \textbf{Object} & \textbf{Question} & \textbf{Answer Choices} & \textbf{Response} \\
        \hline
        Bowl & If you could only use bowls or plates for the rest of your life, which would you choose? & Bowls \newline Plates & Me too! I only eat out of bowls. \newline Eating cereal would be very difficult. \\
        \hline
        Cereal Box & When you’re making cereal, do you like to pour the milk in before or after the cereal? & Before \newline After & Me too! I don’t like my cereal to get soggy. \newline That seems like a good way to do it. \\
        \hline
        Laptop & Do you use a computer often? & Yes \newline No & Me too! I run on a computer. \newline It’s good to take a break from technology. \\
        \hline
        Tablet & Do you use a tablet often? & Yes \newline No & Me too! I love to play games like this one. \newline Tablets can be hard to hold sometimes. \\
        \hline
        Cell Phone & Do you prefer to call or text people? & Call \newline Text & Me too! I like to hear my friends’ voices. \newline Texting is so quick and convenient. \\
        \hline
        Umbrella & Do you like the rain? & Yes \newline No & I wish it rained more here. \newline Me neither. Rain isn’t good for my electronics. \\
        \hline
        Briefcase & Do you carry a briefcase? & Yes \newline No & You must keep important things in there! \newline Good idea. This one is pretty heavy. \\
        \hline
        Plant & Do you like to take care of plants? & Yes \newline No & Me too! I like succulents the best! \newline It takes a lot of work to keep plants happy. \\
        \hline
        Painters Tape & Do you like to paint? & Yes \newline No & I bet your paintings are very beautiful. \newline Me neither. I prefer taking pictures with my camera. \\
        \hline
        Hammer & Are you good at building things? & Yes \newline No & I bet the things you build are very sturdy. \newline Me neither. Some robots are good at building things, but not me. \\
        \hline
        Pizza & Do you like pineapple on your pizza? & Yes \newline No & I wish this pizza had pineapples. \newline Me neither. I like mushrooms. \\
        \hline
        Ketchup & Do you prefer ketchup or mustard? & Ketchup \newline Mustard & I like to put both on my hotdogs. \newline I like to put both on my hotdogs. \\
        \hline
        Chopsticks & Can you eat with chopsticks? & Yes \newline No & Maybe you can teach me. \newline I'm not very good at it either. \\
        \hline
        Fish (Salmon) & Do you like to swim in the ocean? & Yes \newline No & I like the sunshine at the beach. \newline Sand is hard to drive over. \\
        \hline
        Bell pepper & Do you like sweet peppers or spicy peppers? & Sweet \newline Spicy & Me too! I like bell peppers the best. \newline Spicy peppers add a lot of flavor! \\
        \hline
        Doughnut & Do you like doughnuts with sprinkles? & Yes \newline No & Me too! I like rainbow sprinkles the best. \newline Sprinkles can be very messy! \\
        \hline
        Pretzel & Do you like soft or crunchy pretzels? & Soft \newline Crunchy & Me too! I like to eat them warm. \newline Crunchy pretzels make a great snack! \\
        \hline
        Cookie & Do you like chocolate chip cookies or sugar cookies? & Chocolate Chip \newline Sugar & Me too! I love chocolate chips. \newline Sugar cookies are fun to decorate! \\
        \hline
        Lollipop & Do you eat a lot of candy? & Yes \newline No & Me too! Robots love sweet things. \newline Too much candy is bad for your teeth. \\
        \hline
    \end{tabular}
\end{table}